\documentclass[conference]{IEEEtran}
\IEEEoverridecommandlockouts
\usepackage{cite}
\usepackage{amsmath,amssymb,amsfonts}
\usepackage{algorithmic}
\usepackage{graphicx}
\usepackage{textcomp}
\usepackage{xcolor}
\usepackage{tabularx}
    \newcolumntype{L}{>{\raggedright\arraybackslash}X}
\usepackage{siunitx}
\usepackage{upgreek}
\usepackage{xurl}
\usepackage{xr}
\makeatletter
\newcommand*{\addFileDependency}[1]{
  \typeout{(#1)}
  \@addtofilelist{#1}
  \IfFileExists{#1}{}{\typeout{No file #1.}}
}
\makeatother

\usepackage{cleveref}
\usepackage{soul}

\usepackage[symbol]{footmisc}

\usepackage{listings}
\usepackage{xcolor}

\def\BibTeX{{\rm B\kern-.05em{\sc i\kern-.025em b}\kern-.08em
    T\kern-.1667em\lower.7ex\hbox{E}\kern-.125emX}}
\begin{document}

\title{Answer Fast: Accelerating BERT on the Tensor Streaming Processor
}

\author{\IEEEauthorblockN{Ibrahim Ahmed, Sahil Parmar, Matthew Boyd, Michael Beidler, \\ Kris Kang, Bill Liu, Kyle Roach, John Kim and Dennis Abts
}
\IEEEauthorblockA{Groq Inc. \\ \{iahmed, sparmar, matt, mb, kkang, bliu, kroach, jkim, dabts\}@groq.com}

}

\maketitle

\begin{abstract}

Transformers have become a predominant machine learning workload, they are not only the de-facto standard for natural language processing tasks, but they are also being deployed in other domains such as vision and speech recognition. Many of the transformer-based applications are real-time systems such as machine translation and web search. These real time systems often come with strict end-to-end inference latency requirements. Unfortunately, while the majority of the transformer computation comes from matrix multiplications, transformers also include several non-linear components that tend to become the bottleneck during an inference. In this work, we accelerate the inference of BERT models on the tensor streaming processor. By carefully fusing all the nonlinear components with the matrix multiplication components, we are able to efficiently utilize the on-chip matrix multiplication units resulting in a deterministic tail latency of 130 \textmu s for a batch-1 inference through BERT-base, which is 6$\times$ faster than the current state-of-the-art.
\end{abstract}

\section{Introduction}

Transformer-based models~\cite{transformer_paper} have revolutionized various natural language processing (NLP) applications; state-of-the-art results in machine translation~\cite{translation}, web search~\cite{google_bert}, question and answering~\cite{SQUAD_paper_2} almost exclusively use transformers. In addition to dominating the NLP domain, transformers are starting to penetrate other domains such as computer vision~\cite{vit_paper} and speech recognition~\cite{speech_transformer}.  Many of the production-level transformer-based models are real-time systems where users interact with a service and expect a response in real-time, which enforces a strict latency requirement during inference. 

BERT~\cite{BERT_paper} is a popular transformer model that is widely used in the industry: Microsoft~\cite{microsoft_bert} and Google~\cite{google_bert} search engines rely on BERT models; Twitter~\cite{twitter_bert} content moderation pipeline also includes a BERT model; and Roblox~\cite{roblox_bert} uses BERT as part of their text classification and named entity recognition pipelines. In many services, an inference through the BERT model is usually one component that feeds other downstream tasks before returning an answer to the user. As such, to guarantee a reasonable service time it is important to ensure that the observed (average) latency and {\em tail} latency of the inference is under the strict latency budget available to the entire pipeline. In addition, any reduction in the inference latency would relax the timing constraints for the other components in the rest of the pipeline.

The increasing demand for compute by machine learning applications has resulted in a large number of domain-specific accelerator chips.  While most of these new chips share the objective of maximizing their compute capabilities per silicon area, preliminary results suggest that each chip has a unique advantage. For example: Cerebras Wafer Scale Engine has shown that it is able to accelerate training large-scale models~\cite{cerebras_paper} compared to GPUs, Tenstorrent's Grayskull focuses on conditional computing (the ability to dynamically prune parts of the model based on the input)~\cite{linley_tenstorrent}, and Groq's deterministic Tensor Streaming Processor (TSP) has shown promising results for batch-1 inference~\cite{thinkfast}.


The goal of this work is to accelerate  a BERT model inference with the 
objective of minimizing both {\em average} latency and latency {\em variation} (including tail latency). We exploit the Groq TSP \cite{thinkfast} hardware accelerator that provides high batch-1 performance while supporting deterministic execution to minimize latency variation. Our work shows that predictable performance can be achieved, compared to a modern GPU, while also achieving significantly lower average latency.   In particular, the contribution of this work includes the following:

\begin{itemize}
\item give an overview of the TSP microarchitecture and programming model, 
\item efficiently map the computation of BERT to the TSP's vector and matrix execution units and its streaming architecture,
\item describe the implementation of the BERT model using the GroqAPI framework provided as part of Groq software developer kit (SDK), and
\item characterize the model's performance on thousands of inferences by measuring the observed end-to-end latency and distribution of response times. 
\end{itemize}

The remainder of this paper is organized to describe the BERT model architecture and its functional mapping and implementation using the GroqAPI programming framework. To show statistically significant results, we perform thousands of inferences on the TSP to measure its response latency and distribution of response times. We compare our approach to other related work and summarize our results by showing a 6$\times$ reduction in execution time compared to the same model on a current-generation GPU while providing deterministic latency and minimizing tail latency.   
 


\definecolor{codegreen}{rgb}{0,0.6,0}
\definecolor{codegray}{rgb}{0.5,0.5,0.5}
\definecolor{codepurple}{rgb}{0.58,0,0.82}
\definecolor{backcolour}{rgb}{0.95,0.95,0.92}
\lstdefinestyle{mystyle}{
  backgroundcolor=\color{backcolour}, commentstyle=\color{codegreen},
  keywordstyle=\color{magenta},
  numberstyle=\tiny\color{codegray},
  stringstyle=\color{codepurple},
  basicstyle=\ttfamily\footnotesize,
  breakatwhitespace=false,         
  breaklines=true,                 
  captionpos=b,                    
  keepspaces=true,                 
  numbers=left,                    
  numbersep=5pt,                  
  showspaces=false,                
  showstringspaces=false,
  showtabs=false,                  
  tabsize=2
}
\lstset{style=mystyle}

\section{Background}
\subsection{BERT}
BERT~\cite{BERT_paper} architecture is almost identical to the encoder stack component of the original transformer paper~\cite{transformer_paper}. It is composed of an embedding layer, an encoder stack and an output layer. A pre-trained BERT model can be fine-tuned to target specific tasks (e.g. Question Answering) by changing the output layer to match the down-stream tasks during the fine-tuning step. The encoder stack is built from $N$ identical layers (where $N$ is a model hyper-parameter); each layer has a multi-headed self-attention block and a feed-forward block. 
Since the focus of this work is to accelerate BERT, we will discuss the different computations involved in an encoder layer and we will refer the reader to the BERT paper~\cite{BERT_paper} for a better understanding of the intuition behind these computations.  The computations involved in a multi-headed self-attention block in each encoder layer of BERT~\cite{BERT_paper} are listed below:

\begin{equation}\label{Q}
Q_{i} = XW_{i}^{q}+b_{i}^{q}.
\end{equation}
\begin{equation}\label{K}
K_{i} = XW_{i}^{k}+b_{i}^{k}
\end{equation}
\begin{equation}\label{V}
V_{i} = XW_{i}^{v}+b_{i}^{v}
\end{equation}
\begin{equation}\label{qk_softmax}
head^{i} = \mathrm{softmax}(\frac{Q_{i}K_{i}^{T}}{\sqrt{d_{k}}})V_{i}
\end{equation}
\begin{equation}\label{sa_out}
Sa = \mathrm{LN}\footnote{LN: layernorm operation}(Concat(head_{1},...,head_{h})W^{0}+b^{0}+X)
\end{equation}
where $X$ and $Sa$ are the input and output of the self-attention block, respectively. ${W_{i}^{*}}$ and ${b_{i}^{*}}$ are the model weights and biases, while $h$ and $d_{k}$ are hyper-parameters representing the number of heads and the head size, respectively. The output of the self-attention block is passed to a feed-forward layer that performs the following:
\begin{equation}\label{enc_layer_out}
layer\_out = \mathrm{LN}(\mathrm{GELU}(SaW^{1}+b^{1})W^{2}+b^{2}+Sa)
\end{equation}
where GELU is the Guassian Error Linear Unit~\cite{gelu}. The output ($layer\_out$) of a layer feeds the next layer in the encoder stack, and the embedding layer feeds the first encoder layer.

\begin{figure}
    \centering
    \includegraphics[width=0.7\linewidth]{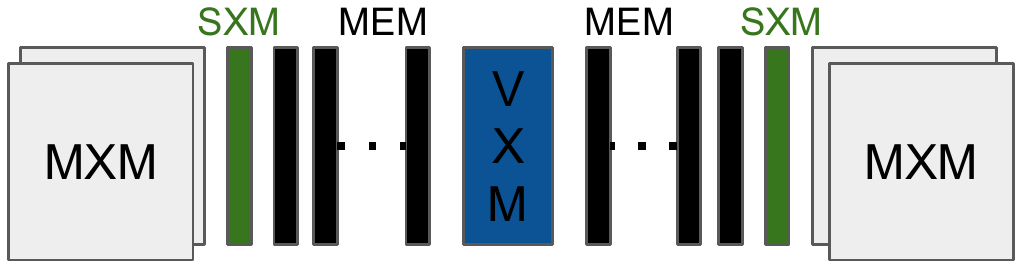}
    \caption{Simplified view of parts of the TSP architecture.}
    \label{tsp_Arch}
\end{figure}

\subsection{TSP Architecture}
The Tensor Streaming Processor (TSP), a statically scheduled SIMD architecture, performs computation using a streaming processing model where computational elements, arranged spatially by function, perform operations as tensor data streams over the functional units. 
The architecture positions computation and memory units horizontally across the chip taking advantage of dataflow locality to reduce latency and power and increase communication bandwidth. 
The TSP uses a tiled microarchitecture to scale the vector length to the underlying data up to 320 bytes in length. 

The high-level microarchitecture and spatial layout of the TSP architecture~\cite{thinkfast} is shown in Fig.~\ref{tsp_Arch}. 
Matrix execution modules (MXMs) are located on the East and West sides of the chip. 
Switching execution modules (SXMs) are located inside of the MXMs. 
The Vector execution module (VXM) is located in the center. 
SRAM memory modules (MEM) are located between the computational units providing localized, high bandwidth, low latency, load and store. 

Data is streamed between functional units through 45 streaming register files (SRFs) positioned across the chip. 64 streams (32 Eastward and 32 Westward) continuously move data between SRFs, each carrying a 320-byte vector, one SRF per cycle, at 20 TiB/sec. Functional units consume inputs and produce outputs from and to sets of streams on adjacent SRFs. With spatial locality between producing and consuming functional units multiple producer/consumer pairs can utilize the same set of streams without data conflicts. 
MEM is composed of 88 independent on-chip memory blocks (distributed evenly between hemispheres). Each memory block contains pseudo-dual-port SRAM cells that support simultaneous read and write operations on opposite banks. Each read or write operation produces or consumes a 320-byte vector per cycle. The on-chip SRAM bandwidth is 55 TiB/sec. 

The TSP consists of four MXM planes; 320$\times$320 2D MACC arrays\footnote{MXM supports int8 and fp16 matrix multiplication\cite{thinkfast}}. Each of the 4 MXM planes is able to independently compute a matrix multiplication between an installed int8 320$\times$320 matrix and a continuously streamed int8 320$\times$N matrix producing a 320-value int32 vector each cycle. 
A physical vector on the TSP can have a maximum length of 320 elements; logical vectors of larger sizes are decomposed into several physical vectors. 

The VXM contains 16 ALUs responsible for performing point-wise arithmetic (add, multiply, tanh, etc.). Each ALU operates on 1 to 8 input operands and produces 1 to 4 outputs. The VXM supports numeric data types from int8 up to fp32. ALU operation chain together to form complex computation pipelines without the need to store intermediate results to memory. The 40 TiB/s at the boundary between VXM and MEM enables full utilization of the VXM compute capability. 

The SXM is a heterogeneous group of units that distribute, mask, transpose, rotate, permute, and shift data within and across vectors. With the SXM located between MEM and MXM is has locality to to MEM for high bandwidth, up to 16 concurrent stream data flow operations and conveniently chains between MEM and the MXM when transforming weights or inputs for matrix multiply operations. The distribute, transpose, and permute and shift units allow chaining within the SXM to avoid writing intermediate results to SRAM for compound data flow transformations.

\subsection{GroqAPI}
In order to leverage the strength of the architecture and maintain full control of our implementation strategy we used GroqAPI to implement BERT. GroqAPI mixes high and low levels of programming abstraction to maintain complete control over the TSP architecture's resources and instruction schedule. GroqAPI allows us to statically schedule all operations on the TSP as we desire. We also have control over how to allocate resources such as VXM ALUs and MXM planes, and how to allocate memory.


We were able to design our compilation strategy for the architecture, predict program latency, implement BERT according to our design, and achieve program latency with 1\% of our original predicted latency. 






\section{Accelerating BERT}

We focus on optimizing a single layer of the encoder stack as it is the fundamental building block of transformers, 
and it makes up most of the inference latency since the encoder is built by repeating this layer multiple times.
The computations involved in that layer (\cref{K,Q,V,qk_softmax,sa_out,enc_layer_out}) are mainly dominated by matrix-matrix multiplications which makes it amenable to acceleration on chips with dedicated matrix multiplication units. 
Unfortunately, the presence of the various non-linear components (softmax, layernorm and GELU) usually results in under utilizing the matrix multiplication units as they have to stay idle waiting for results from these layers.
To accelerate BERT on the TSP, we leveraged the chaining capability of the VXM to pipeline the nonlinear computations with matrix multiplications such that we maximize the utilization of the MXM units on the TSP. 

In this work we use a mixed precision approach for a quantized BERT\footnote{Significantly reduces model size with minimal impact on accuracy\cite{q8bert}.}. 
We target int8 operands for matrix multiplications and fp32 for all non-linear components.
\begin{figure}
    \centering
    \includegraphics[width=0.7\linewidth]{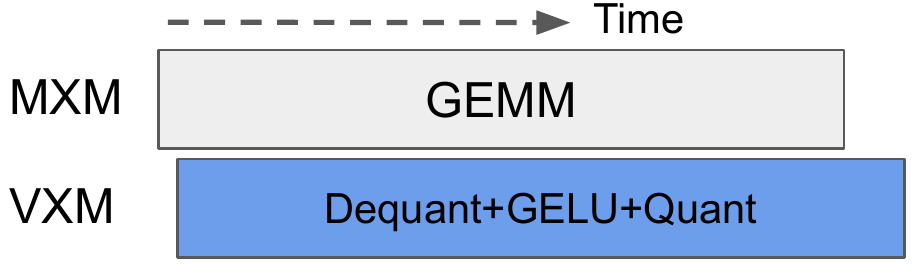}
    \caption{GEMM and GELU overlapping execution time. Grey and blue are performed on MXM and VXM, respectively.}
    \label{gemm_gelu}
\end{figure}

\subsection{GELU}
The input to the non-linear GELU function is the output of a general matrix multiply (GEMM) in the form of $AW + b$ as shown in \cref{enc_layer_out}. GEMMs can be directly mapped to the MXM and the associated accumulator. Since each MXM plane has 320x320 MACCs, if $W$ is larger than that, the multiplication will involve multiple passes (loading weights and then streaming activation). After computing the GEMM, we need to compute GELU, which can be approximated as follows:
\begin{equation}\label{gelu}
GELU(x) = 0.5x(1+tanh(\sqrt{\frac{2}{\pi}}(x+0.044715x^3))) 
\end{equation}
We map GELU onto the VXM by building a pipelined chain of 13 ALUs. 
This pipelined chain produces a new result vector every clock cycle. 
The int32 GEMM result needs to be dequantized~\cite{nvidia_quant} to fp32 before calculating GELU, and the output of GELU needs to be quantized back to int8 before being consumed by downstream GEMMs. 
We use the remaining 3 ALUs of the VXM to pipeline the dequantization and quantization stages with GELU. 
Since downstream GEMMs have to be scheduled after GELU starts generating results, if we scheduled GELU to execute after the upstream GEMM has finished computing, we would keep the MXM idle for the execution time of GELU. 
The input to GELU is the largest activation tensor in BERT, and with a throughput of one result vector per clock cycle the execution time of GELU is equal to or larger than that of the GEMM feeding it\footnote{This is true for all the standard sizes of BERT.}. 
To reduce the MXM idle time, we pipeline the GEMM with GELU such that every output vector from the GEMM is directly sent to the VXM to start computing GELU. 
As shown in Fig.~\ref{gemm_gelu} with this pipelining approach, we effectively hide most of GELU's latency behind the GEMM execution.

\subsection{Layer Normalization}
As seen in \cref{sa_out,enc_layer_out}, the self-attention and feed-forward blocks perform layer normalization (LN). Both LN operations have the form $\mathrm{LN}(\mathrm{dequantize}(X) + Y)$ where $X$ is the int32 output tensor of a GEMM and $Y$ is the fp32 output tensor of a previous LN. Layer normalization is calculated~\cite{layernorm_paper} as follows:

\begin{equation}\label{layernorm_eq}
\mathrm{LN}(Z) = \frac{Z-\mathrm{E(}Z\mathrm{)}}{\sqrt{{\mathrm{VAR(}Z\mathrm{)}}+\epsilon}}\gamma + \beta
\end{equation}

where E($Z$) and VAR($Z$) are the mean and variance of tensor $Z$ along the inner dimension, respectively. If $Z$ has the shape of $(k, j)$, then E($Z$) and {VAR($Z$)} will have the shape of $(k, 1)$, which will be broadcast back to (${k, j}$) when performing point-wise operations with ${Z}$. $\gamma$ and $\beta$ are learnable parameters and $\epsilon$ is a small value to avoid any potential division by zero. Performing layer normalization requires three sequential passes over the input tensor ${Z}$: one pass to calculate E($Z$), a pass to calculate VAR(${Z}$), and a final pass to normalize ${Z}$. We leverage all 16 ALUs of the VXM to accelerate these passes.

\begin{figure}
    \centering
    \includegraphics[width=0.8\linewidth]{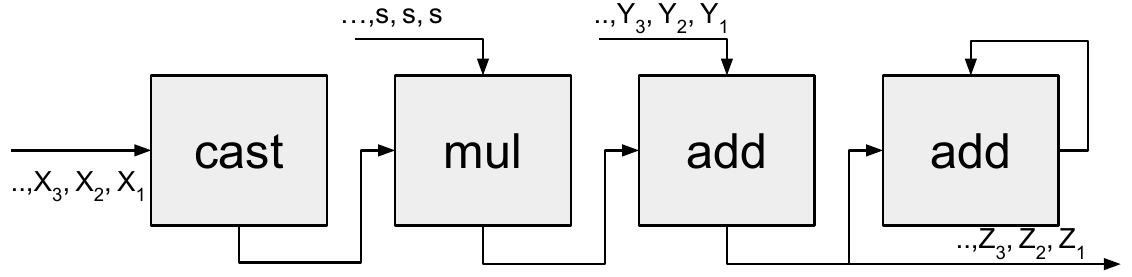}
    \caption{A single chain of 4 ALUs performing the first pass of LN.}
    \label{ln_1_pipe}
\end{figure}

\subsubsection{First Pass}
To start the LN, we need to first compute Z, and so we overlap producing $Z$ with calculating E($Z$). As shown in Fig.~\ref{ln_1_pipe}, we build a chain of three ALUs that dequantizes $X$ (cast and a multiplication by a constant) and adds the result to $Y$ to calculate $Z$. This chain generates a vector of $Z$ ($Z_{i}$) every cycle, where $Z_i$ represents the i\textsuperscript{th} column of the tensor. Z is transmitted from the VXM to memory, but it is also directly sent to another ALU to sum all the vectors of Z. Since this chain only needs four ALUs, we built four parallel chains so that we generate four vectors of Z every cycle. After producing all vectors of Z, we calculate the average of Z by adding the partial sums (from the parallel chains) and dividing the result by the number of vectors in Z. The outputs of the first pass are $Z$ and E($Z$).

\begin{figure}
    \centering
    \includegraphics[width=0.85\linewidth]{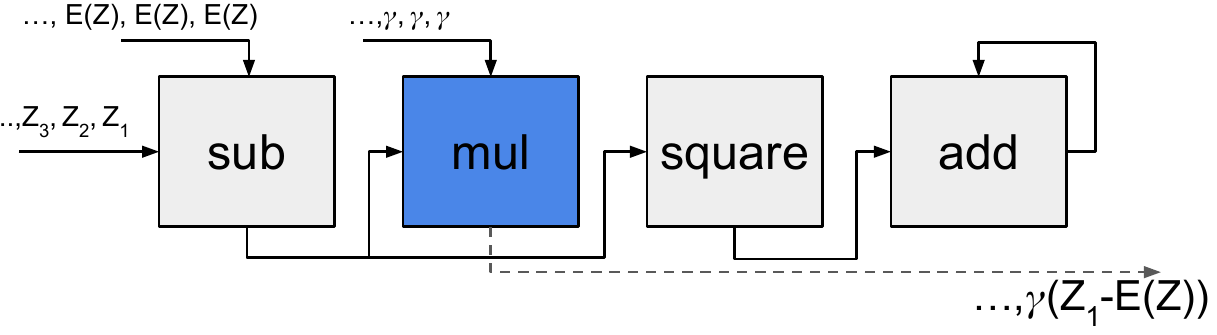}
    \caption{A single chain of 4 ALUs performing the second pass of LN.}
    \label{ln_2_pipe}
\end{figure}
\subsubsection{Second Pass}
After calculating the mean, we need to compute the variance as follows:
\begin{equation}\label{lvar_eq}
\mathrm{VAR}(Z) = \mathrm{E}((Z-\mathrm{E}(Z))^{2})
\end{equation}
Since we have already computed ${Z}$ and E($Z$), we can map \cref{lvar_eq} to a chain of three ALUs as shown by the grey ALUs in Fig.~\ref{ln_2_pipe}: first ALU performs the subtraction, second ALU squares the difference, and a third ALU to accumulate the incoming vectors.

The term $Z-\mathrm{E}(Z)$ is needed in \cref{layernorm_eq} and \cref{lvar_eq}, so we can store that result from the second pass and reuse it later. However, to reduce the number of ALUs used in the third pass, we add another ALU to calculate $\gamma(Z-\mathrm{E}(Z))$. As shown in Fig.~\ref{ln_2_pipe}, the output of this multiplication is performed by the blue ALU and is written to memory. Similarly to the first pass, the 4-ALU chain in the second pass also produces one output vector every cycle, and we create four parallel chains to increase the concurrency to four vectors per cycle.

At the end of the second pass, we add $\epsilon$ to VAR($Z$) and calculate the reciprocal square root of the result. The outputs of the second pass are the numerator and denominator of the first term in~\cref{layernorm_eq}.

\begin{figure}
    \centering
    \includegraphics[width=1\linewidth]{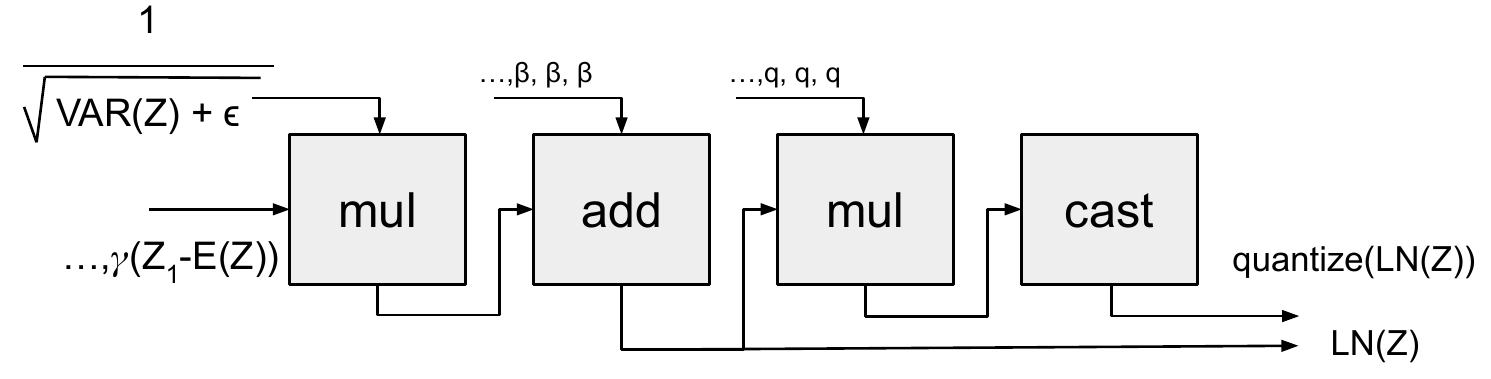}
    \caption{A single chain of 4 ALUs performing the last pass of LN.}
    \label{ln_3_pipe}
\end{figure}

\begin{figure}
    \centering
    \includegraphics[width=0.8\linewidth]{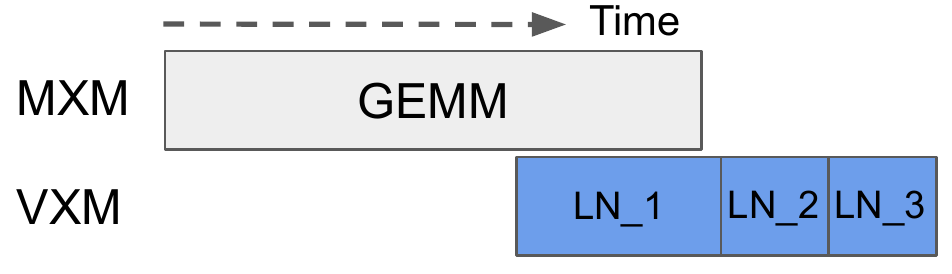}
    \caption{GEMM and LN overlapping execution time. Grey and blue are performed on MXM and VXM, respectively.}
    \label{gemm_ln}
\end{figure}

\subsubsection{Third Pass}
In an int8 quantized BERT, the output of the layer normalization is consumed twice: it gets multiplied by a weight matrix in a downstream GEMM, and it gets added to another tensor just before being consumed by another layer normalization block. In the final pass of LN, we calculate both an int8 and and fp32 output. Similarly to the previous two passes, we use a chain of four ALUs; as shown in Fig.~\ref{ln_3_pipe} the first ALU calculates the product of the outputs generated in the second pass and the second ALU adds $\beta$ to the result. The output of the second ALU is the fp32 results of the layer normalization block (LN($Z$)). This output is quantized as it is produced to also generate the int8 quantized tensor which will be sent to the downstream GEMM. We also use four parallel chains in this pass.

With our LN implementation, we are able to fully utilize all 16 ALUs of the VXM during the entire execution time. Since we calculate the mean while producing $Z$ and we reuse $Z-\mathrm{E}(Z)$ from the second pass, we only need to read $Z$ from memory once throughout the three LN passes. Assuming the shape of $Z$ is $(k, j)$, the number of cycles needed for the layer normalization is:

\begin{equation}\label{ln_cycles}
LN\_cycles = 3 * j * \lceil \frac{k}{320} \rceil * \frac{1}{4} + c
\end{equation}

In each of the three passes, we have a throughput of 4 physical vectors per cycle. 
Number of physical vectors depends on how many rows ($j$) are in Z and how many physical vectors compose a single column (each physical vector can hold a maximum of 320 elements). 
After the first and second passes, we perform a normalization step that requires a constant number of cycles ($c$), which does not change with the size of $Z$.

As a final optimization, we start executing the first LN pass while the GEMM producing $X$ (one of the input to LN) is executing. 
As shown in Fig.~\ref{gemm_ln}, This optimization effectively hides the latency of the first LN pass completely. 
With this optimization, the MXM unit will only be idle for the time needed to finish the second and third LN passes. 

\begin{figure}
    \centering
    \includegraphics[width=0.9\linewidth]{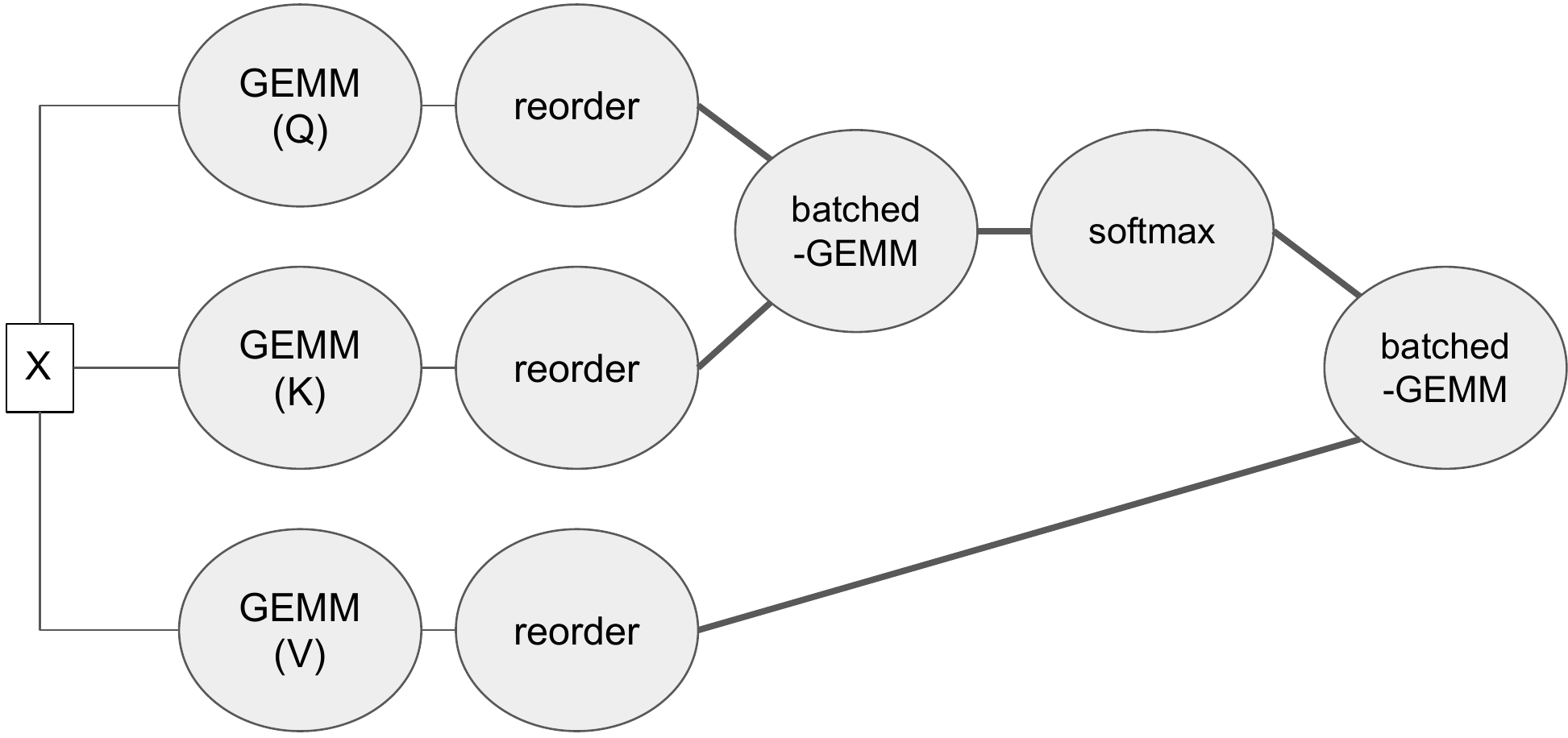}
    \caption{Compute graph for the self-attention block.}
    \label{sa_block_df}
\end{figure}

\subsection{Self-Attention Block}
Within the self-attention block, we need to perform \cref{K,Q,V,qk_softmax} for every head in the model where the number of heads ($h$) is a model parameter\footnote{The standard BERT sizes use 12 and 24 heads.}. 
Instead of performing $h$ separate matrix multiplications to compute $Q_i$ (\cref{Q}), a common transformer optimization is to concatenate the different weights ($W_i^q$) of all the heads and perform one large matrix multiplication that generates ${Q}$ which includes the $Q_i$ of all the heads. 
In our implementation, we follow this optimization strategy for calculating $Q$, $K$ and $V$.

Fig.~\ref{sa_block_df} shows the compute graph of the self-attention block. 
It includes nodes (reorder) representing the reshape and transposition operations needed to separate the different heads from the concatenated $Q$, $K$ and $V$. 
A thick line between two nodes represent a change in data type performed as a quantization or a dequantization step. 
As mentioned earlier, the output of our GEMM is int32, but the inputs are expected to be int8, and the inputs and outputs of the softmax are in fp32. 
The batched-GEMM represents several independent GEMMs (one for each head) to perform the GEMMs in \cref{qk_softmax} for all the heads in the model.

We parallelize the execution of GEMMs across all MXM planes. When a GEMM requires multiple MXM passes (happens when the weights tensor has more than 320 rows or columns) on the same MXM plane, we hide the delay of installing weights by loading the weights of pass $i$ while executing pass $i-1$, so there are no idle MXM cycles between passes; that is, each MXM plane is continuously producing data.

Performing transpositions to separate the different attention heads can be an expensive operation. Since the output of these transpositions are only used as inputs to the batched-GEMM, we simplified this step to avoid transpositions and just rely on reshaping and masking the inputs being sent to the MXM. With this simplification, the reorder operation shown in Fig.~\ref{sa_block_df} is performed on-the-fly in the SXM as data is travelling from memory to the MXM.

The non-linear component in the self-attention block is the softmax operation. Similar to layer normalization, softmax requires more than one pass on the input tensor. However, we need to perform $h$ independent softmax operations (one for each head). We have optimized the softmax in a similar approach to the one used in optimizing the layer normalization; we store intermediate values from one pass that can be reused in another pass, and we build four parallel chains of ALUs that allow us to produce four vectors of results during every pass of the softmax. 

\begin{figure}
    \centering
    \includegraphics[width=0.9\linewidth]{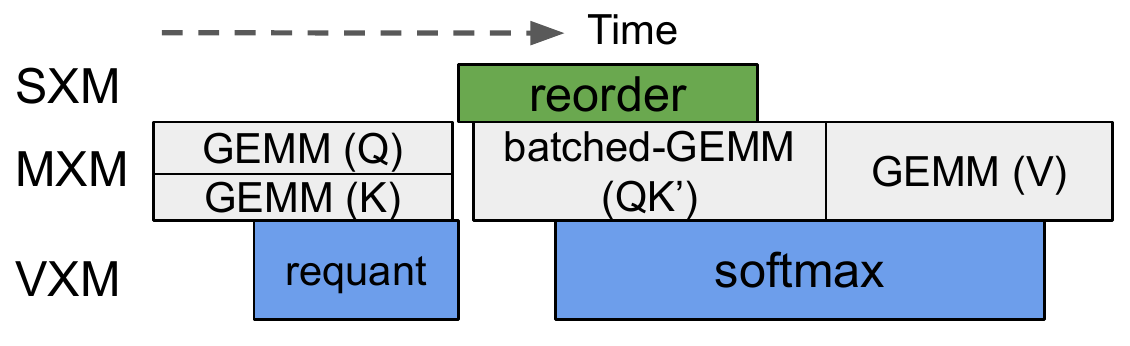}
    \caption{Self-attention block execution schedule. Grey, blue and green are performed on MXM, VXM and SXM, respectively.}
    \label{sa_sched}
\end{figure}

Fig.~\ref{sa_sched} shows the scheduling of the softmax and reorder operations with the other GEMMs in the self-attention block. 
By pipelining the reorder operation with the execution of the batched-GEMM, we can start the batched-GEMM without waiting for the reorder operation to finish execution. 
We were also able to hide the latency of the softmax operation completely by starting its execution as soon as the first vector of results is produced by the batched-GEMM operation, and overlapping the last part of the softmax with another independent GEMM (calculating $V$ does not have a dependency on the result of softmax). 
Note that during the batched-GEMM shown in Fig.~\ref{sa_sched}, we have a pipeline that starts by reading a vector from MEM, reordering it on the SXM, passing it to the MXM, then sending the MXM result to the VXM to flow through several ALUs (softmax pass) to be finally written to MEM again.

The key optimizations that enabled us to achieve low latency on batch-1 inferences are:
\begin{enumerate}
    \item Deeply pipelining non-GEMM operations with GEMMs to hide their latency and increase the utilization of the MXM.
    \item Optimizing the layer normalization operation to reduce the idle time time during which the MXM is waiting for LN results.
\end{enumerate}
This deep pipelining also resulted in reducing the on-chip scratchpad memory needed for intermediate results, which leaves more on-chip memory to be used for constant data.






%

\section{Results}
Using GroqAPI we implemented a quantized BERT-base model to execute on the TSP hardware; our implementation included all the optimization strategies explained earlier. BERT-base models have 12 attention heads ($h$), 12 layers ($N$) in the encoder stack, each head has a size of 64 ($d_k$), and the feed-forward block has a size of 3072. We added an output layer to our model to target question answering tasks. To get the int8 quantized model similar to~\cite{q8bert}, we started with a pretrained BERT base model (uncased)~\cite{bertbase_huggingface}, added fake quantization nodes to the model (introducing quantization errors), calibrated the fake quantization nodes to get the scalar values needed for scaling, and then performed quantization-aware-training (QAT) by fine tuning the quantized model using the SQuAD 1.1 training dataset~\cite{SQuAD_dataset}. The quantization method used was per-tensor, scale-based linear quantization~\cite{nvidia_quant}. We explored two quantized models: one (\textit{quant-uniform}) that has all the biases quantized to int32, and weights and embedding quantized to int8; and another (\textit{quant-mixed}) that has the weights quantized to int8, embedding quantized to int16, and biases quantized to int32. The fp32 baseline F1 score of the model running SQuAD 1.1 dev dataset was 87.18\%, \textit{quant-uniform} achieved an F1 score of 86.98\%, and \textit{quant-mixed} achieved 87.16\%. Unless otherwise stated, for the remainder of this section we will be discussing performance of the \textit{quant-mixed} model.

TSP results shown here were measured on a remote server. The TSP was running at a 900 MHz clock, and the server has 32 CPU cores running an Ubuntu 18.04.4 OS.

\begin{figure}
    \centering
    \includegraphics[width=0.85\linewidth]{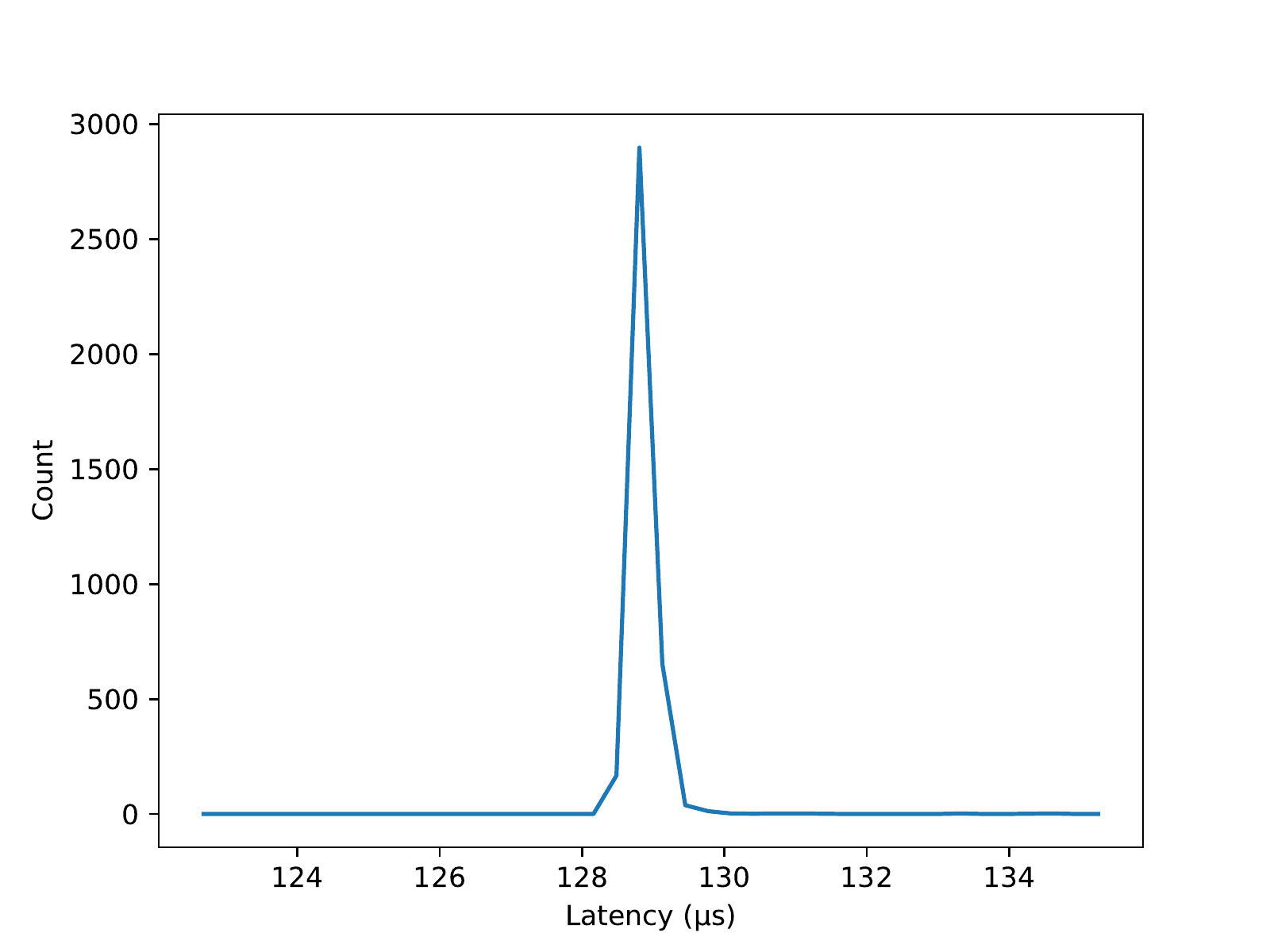}
    \caption{BERT inference latency distribution across 4000 inferences.}
    \label{latency_dist}
\end{figure}

\begin{figure}
    \centering
    \includegraphics[width=0.85\linewidth]{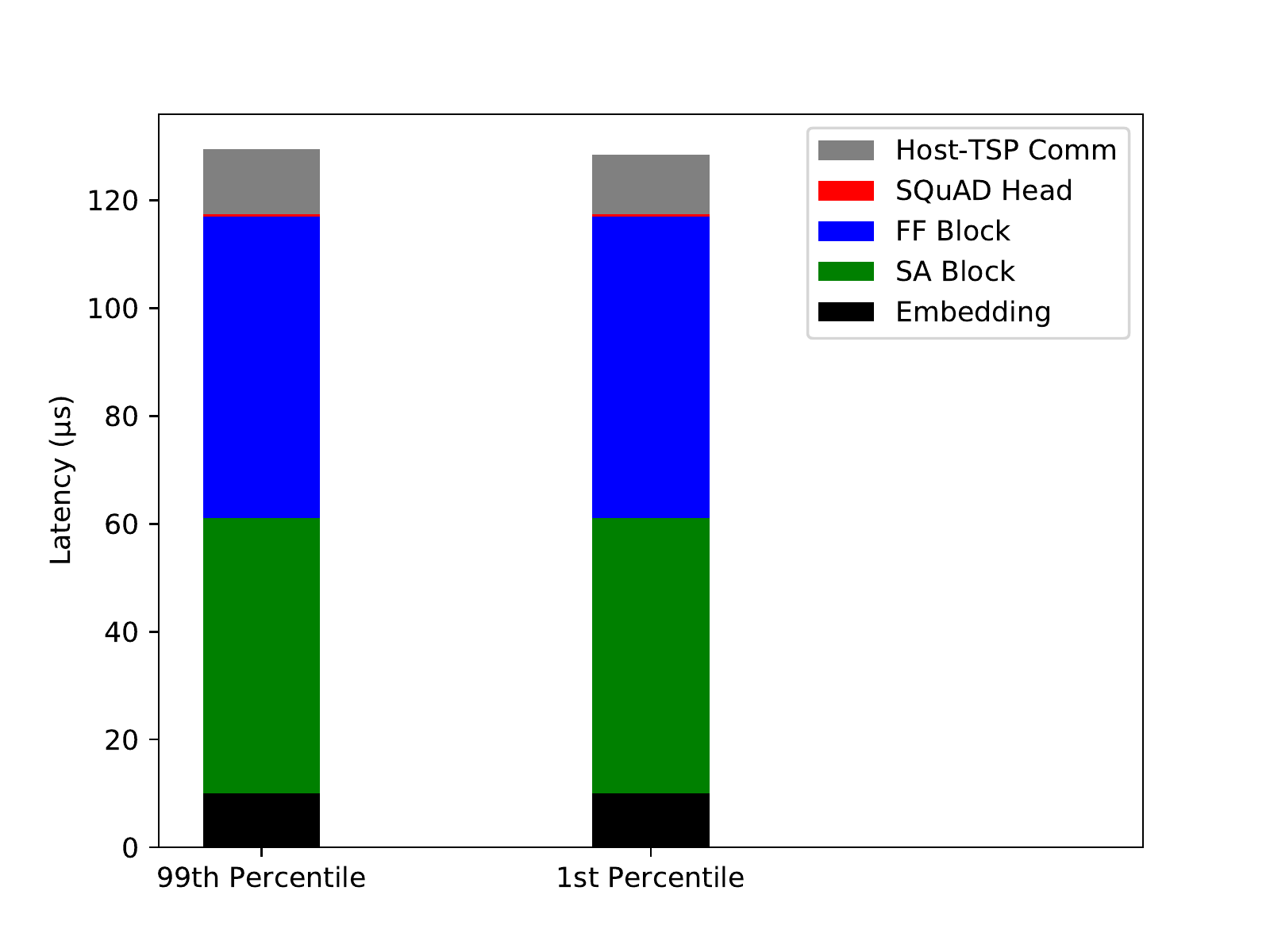}
    \caption{Breakdown of the 99\textsuperscript{th} and the 1\textsuperscript{st} percentile latency.}
    \label{latency_breakdown}
\end{figure}
\subsection{Deterministic Low Latency}
A key advantage of our implementation on the TSP is that it not only provides low latency, but that it provides low tail latency. Since the TSP is deterministic, we have full control in scheduling instructions and so our implementation does not suffer from variation in latency due to hardware-managed scheduling or non-uniform memory access latency. In fact, as mentioned earlier, before implementing our mapping of BERT in GroqAPI we estimated the latency of that mapping and it was only 1\% longer than what we eventually measured on hardware. Having full control over scheduling every instruction enabled us to achieve the deep pipelining needed to hide the latency of all non-GEMM components to maximize the utilization of the MXM unit. 

We measured the latency of thousands of invocations of our BERT model running on the TSP; this latency includes the time to copy data from a host to the TSP, time to execute the model on the TSP, and time to copy results back to the host. Our measured latency does not include the tokenization time of the input sequence and the time to project the final prediction of the model back to text. Tokenization~\cite{tokenization_paper} is the process of translating the input text into a list of tokens that will be consumed by the model. Tokenization and final projection are not compute-heavy and so we perform them on the host\footnote{This is a standard practice when accelerating NLP models~\cite{nvidia_bert}.}. We are focused on batch-1 execution, so every invocation performs an inference on a single input sequence; we consistently used inputs with 128 sequence length. Fig.~\ref{latency_dist} shows the distribution of the latency for 4000 different inferences. As expected, the deterministic execution resulted in a very narrow latency distribution with an average latency of 128.9 \textmu s and a standard deviation of only 3.8 \textmu s.

To better understand the distribution of end-to-end response times, we decomposed the execution time of the measured inferences into five chunks:
time needed to execute the embedding layer (Embedding), time spent executing the self-attention blocks of all layers in the encoder (SA Block), time spent executing the feed-forward blocks of all layers in the encoder (FF Block), time to execute the output layer (SQuAD Head), and communication time between host and TSP (Host-TSP Comm). Fig \ref{latency_breakdown} shows that 83\% of the end-to-end latency is from the encoder stack (FF Block and SA Block); this time is divided equally between all the layers. This observation highlights the importance of having a highly-optimized layer in order to reduce the overall inference latency. Fig \ref{latency_breakdown} also shows that the 1\textsuperscript{st} and the 99\textsuperscript{th} percentile latency are 128.5 \textmu s and 129.5 \textmu s, respectively. Both have an identical breakdown with the exception of having different times for the Host-TSP communication. Data is communicated between the host and TSP through a 16-lane PCIe gen4 channel, which will replay a transmission error across the PCIe link if we receive a corrupted packet. This replay mechanism does introduce a small variability in the observed end-to-end latency, as such this chunk is the only source of variation in latency between inferences.

\begin{table}[t]\caption{A100~\cite{nvidia_a100}, T4~\cite{nvidia_t4} and TSP~\cite{linley_tsp, thinkfast} specifications.}
\begin{center}
\begin{tabularx}{\linewidth}{|L|L|L|L|L|L|}
\hline
Chip & Die Area (mm\textsuperscript{2}) & Tech Process (nm) & Transistor count (B) & TDP (W) \\
\hline
NVIDIA T4 & 545 & 12 & 13.6 & 70\\
\hline
NVIDIA A100 & 826 & 7 & 54.2 & 400\\
\hline
Groq TSP & 725 & 14 & 26.8 & 275\\
\hline
\end{tabularx}
\label{spec_comp}
\end{center}
\end{table}

\begin{table}[t]\caption{T4, Current SOTA (A100)~\cite{nvidia_bert} and this work BERT-base latency (quantized int8 model and 128 sequence length).}
\begin{center}
\begin{tabularx}{\linewidth}{|L|L|L|L|L|}
\hline
 & T4 (\textmu s) & Current SOTA A100 (\textmu s) & This work (\textmu s) & Speedup (This work vs SOTA)\\
\hline
Average & 1330 & 630 & 128.9 & 4.8$\times$\\
\hline
95\textsuperscript{th} Percentile & 1550 & 780 & 129.1 & 6$\times$\\
\hline
99\textsuperscript{th} Percentile & 1570 & 790 & 129.5 & 6.1$\times$\\
\hline
\end{tabularx}
\label{latency_comp}
\end{center}
\end{table}

\subsection{Comparison to State-of-the-art}
Since GPUs dominate the high-performance machine learning inference and training markets, we compare our implementation against the current state-of-the-art (SOTA) int8 quantized BERT-base implemented using the highly-optimized TensorRT and running on the latest A100 GPU~\cite{nvidia_bert}. Table~\ref{spec_comp} compares the specifications of the A100 to the TSP. The A100 is a more recent chip manufactured using a smaller technology node
and has approximately twice the number of transistors compared to the first-generation TSP that we are using.
Table~\ref{spec_comp} also shows the specification for the slightly older T4 GPU that targets markets with lower power profile as evident by the lower TDP and commonly used for inference. 

Table~\ref{latency_comp} lists the latency of NVIDIA-optimized BERT-base running on a T4 GPU, A100 GPU and our implementation running on the TSP. It shows that the A100 sets the record as the current SOTA with an impressive  2$\times$ speedup compared to T4. Similar to our methodology, the A100 latency numbers reported in this table also do not include the time for tokenization and final projection. However, the A100 latency numbers also do not include the time to copy data between host and TSP (while our latency numbers take that into account). The table shows that our implementation achieves a 4.8$\times$ speedup compared the average latency of the current SOTA. Although the A100 has double the number of transistors, our implementation was able to squeeze significantly more effective computing power from the TSP. The TSP architecture enabled us to spatially parallelize independent components and to deeply pipeline components with data dependency such that we hide the latency of non-GEMM compute.

As explained earlier, for real-time applications it is critical to maintain a low tail latency, so we also compare the 95\textsuperscript{th} and 99\textsuperscript{th} percentile latency. The 99\textsuperscript{th} percentile of the current SOTA jumps by more than 25\% compared to the average latency (from 630 \textmu s to 790 \textmu s). This jump is mainly due to contention in the cache-based memory hierarchy used by GPUs. This hierarchy results in a non-uniform latency when accessing memory which translates to a longer tail latency for some inferences. The TSP's on-chip SRAM is explicitly managed by software, and its 220 MiByte capacity is 4-10$\times$ the size of a modern GPU's last-level cache (LLC). Due to the deterministic nature of the TSP, the 99\textsuperscript{th} percentile latency of our implementation is only 0.4\% higher than the average. For real-time applications that care about tail latency, our implementation offers a 6.1$\times$ speedup compared to best-known results run on a complex modern GPU having twice the transistor count.

\begin{figure}
    \centering
    \includegraphics[width=0.75\linewidth]{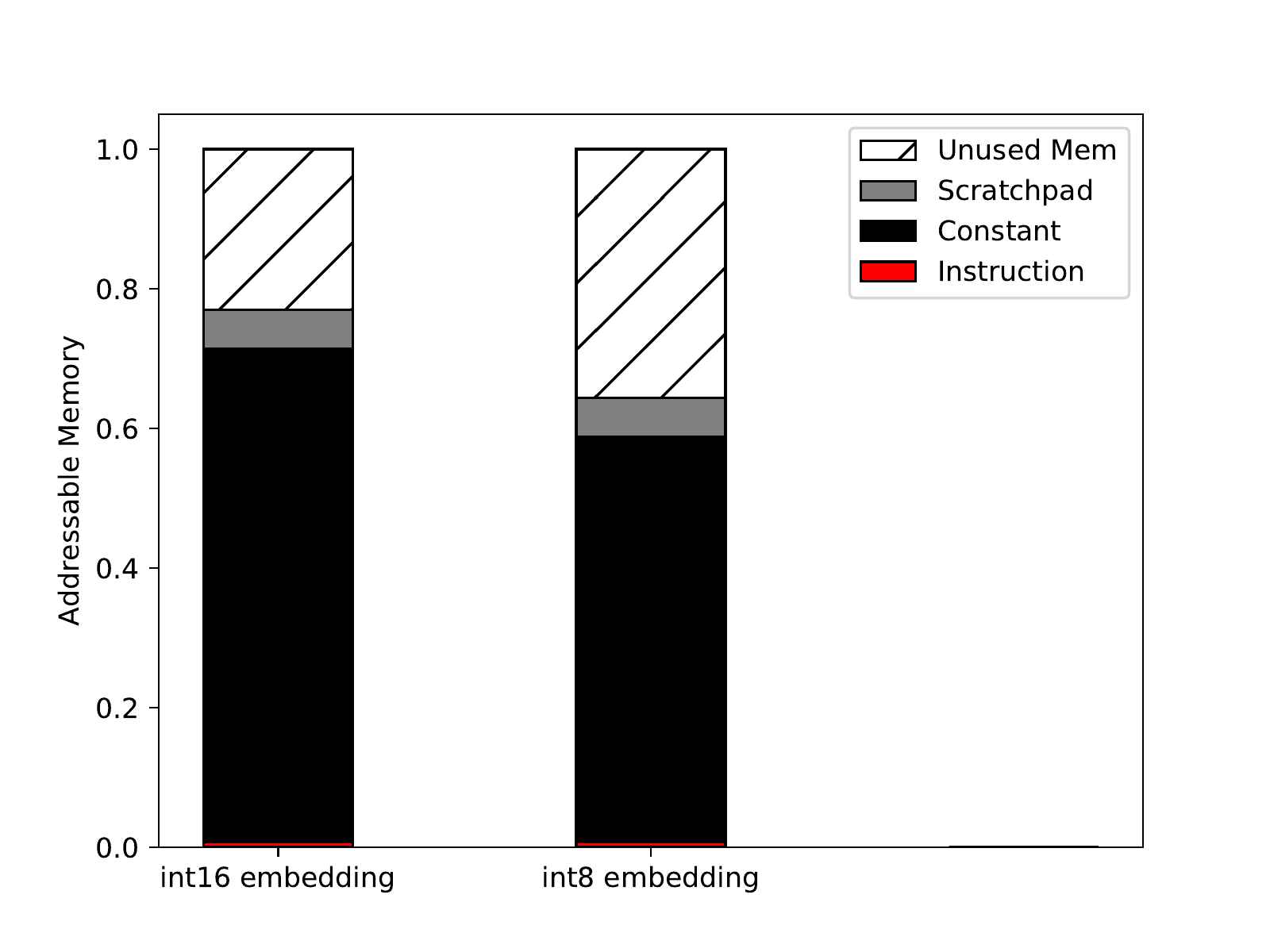}
    \caption{BERT-base TSP memory utilization for different types of embedding.}
    \label{memory_breakdown}
\end{figure}

\subsection{On-chip Memory Analysis}
The TSP's on-chip SRAM capacity of 220 MiBytes is highly ``banked'' to exploit memory-level parallelism. Each memory "slice" contains a bank of SRAM, with up to 176-way memory concurrency. We use this memory concurrency to feed 20 TB/sec of operand bandwidth into the MXM and VXM units and sink their results back to available on-chip SRAM. Since SRAM is more capacity constrained than most DRAM systems, we focused on building BERT models that reside entirely on a single chip. Our implementation of BERT efficiently use the on-chip SRAM by minimizing the amount of memory needed as a scratchpad to store intermediate results. By reducing the size of the scratchpad, we have more on-chip SRAM available to store the constants of the model. Fig.~\ref{memory_breakdown} shows the utilization of the on-chip addressable memory for our implementation of the \textit{quant-uniform} (int8 embedding) and \textit{quant-mixed} (int16 embedding) models. Since our implementation is deeply pipelined, most intermediate results are not written completely to memory but are rather consumed directly by downstream operations or are partially stored in shallow FIFOs. This deep pipelining results in our implementation only consuming $\approx$5.5\% of the addressable memory as a scratchpad. The \textit{quant-mixed} and \textit{quant-uniform} implementations use 71\% and 58\% of the addressable space as constant memory, respectively. The increase in constant memory is due to using a larger embedding for the \textit{quant-mixed} model. Both models use less than 1\% of the addressable memory space to store instructions.

\section{Related Work}
Transformer models, like BERT, are increasingly finding new use-cases in image recognition and natural language processing (NLP) such as {\em question answering} where end-to-end latency is vital to return both a timely and  quality response.
PoWERBERT \cite{pmlr-v119-goyal20a} exploits redundancy in word-vectors and dropping less significant word-vectors; they show a 4$\times$ latency reduction. The techniques they advocate result in $<$1\% accuracy loss, and would equally apply to our approach on the TSP. DeeBERT \cite{xin2020deebert} uses dynamic early exit, similar to that used in vision models, yielding 40\% reduction in inference time with minimal accuracy loss. Similar conditional-exit could be added to our implementation on the TSP to show similar benefit.

Microsoft~\cite{DeepSpeed} accelerated inference of large transformers using multi-GPU systems by efficiently distributing the work load across the different GPUs in the system. They report up to 4.4$\times$ reduction in latency. In our future work, we will also explore accelerating larger models across a system of TSPs. Fang et al.~\cite{turbo_transformer} accelerated different variants of BERT models on GPUs by parallelizing the reduction operations (softmax and layer normalization) and adapting the memory manager to handle different sequence lengths.

A${\textsuperscript{3}}$~\cite{a3} and SpAtten~\cite{spatten} are examples of academic efforts that design custom chips to efficiently accelerate transformer-based models.
Both efforts perform further pruning and approximation on the models along with customizing a hardware architecture that result in large speedups compared to general-purpose chips. Unlike these efforts, in this work we accelerate BERT (without custom approximation or pruning) on a general-purpose ML accelerator chip.
\section{Conclusion and Future Work}
Motivated by the importance of low tail-latency for real-time machine learning systems, we accelerated the inference of a transformer-based model (BERT) on the tensor streaming processor (TSP). We analyzed the compute involved in an inference through BERT and identified that in many cases the non-linear components become a bottleneck that results in underutilizing the matrix multiplication unit. We carefully mapped and scheduled the compute blocks of BERT to the different functional units of the TSP such that we hide the latency of these non-linear components behind matrix multiplications. Our optimized implementation resulted in reducing the end-to-end latency of a BERT-base inference to 130 \textmu s which is 6$\times$ lower than that of the current SOTA achieved on an A100 GPU.

In our next steps, we will increase the size of the BERT encoder layers to fully utilize the TSPs vector length. We will experiment with adding volume to weights to potentially improve model accuracy at the expense of a relatively small increase in latency. Our custom-sized BERT-base will have a head size ($d_k$) of 80, a hidden dimension of 960, and a feed-forward block size of 3840. These new sizes increase the model memory size by $\approx$50\%, but our preliminary analysis show that it will only increase the latency by $<$20\% and only increase the used addressable memory by 24\%. 

\bibliographystyle{IEEEtran}
\fontsize{10pt}{10pt}\selectfont
\bibliography{ref}
\end{document}